\documentclass[10pt,twocolumn,letterpaper]{article}

\usepackage{cvpr}              

\PassOptionsToPackage{table}{xcolor}
\definecolor{cvprblue}{rgb}{0.21,0.49,0.74}
\definecolor{grey}{rgb}{0.5,0.5,0.5}

\usepackage[pagebackref,breaklinks,colorlinks,allcolors=cvprblue]{hyperref}
\usepackage{url}
\usepackage{booktabs}
\usepackage{epsfig}
\usepackage{graphicx}
\usepackage{amsmath}
\usepackage{amssymb}
\usepackage{booktabs}
\usepackage{algorithm}
\usepackage{algpseudocode}
\usepackage{dsfont}
\usepackage{wrapfig}
\usepackage[table]{xcolor}

\usepackage{tikz}
\usepackage{comment}
\usepackage{amsmath,amssymb} 
\usepackage{color}
\usepackage{tabularx}
\usepackage{multirow}
\usepackage{booktabs}
\usepackage{makecell}
\usepackage{enumitem}
\setitemize{noitemsep,topsep=0pt,parsep=0pt,partopsep=0pt}
\usepackage{caption}
\usepackage{bbding}
\usepackage{float}

\def\paperID{14145} 
\def\confName{CVPR}
\def\confYear{2025}

\title{RFMedSAM 2: Automatic Prompt Refinement for Enhanced Volumetric Medical Image Segmentation with SAM 2}

\author{Bin Xie$^1$, Hao Tang$^2$, Yan Yan$^3$, and Gady Agam$^1$ \\ \\ 
$^1$Department of Computer Science, Illinois Institute of Technology, USA \\
$^2$School of Computer Science, Peking University, China \\
$^3$Department of Computer Science, University of Illinois Chicago, USA \\
{\tt\small bxie9@hawk.iit.edu, haotang@pku.edu.cn, yyan55@uic.edu, agam@iit.edu}
}

\begin{document}
\maketitle

\begin{abstract}
Segment Anything Model 2 (SAM 2), a prompt-driven foundation model extending SAM to both image and video domains, has shown superior zero-shot performance compared to its predecessor. Building on SAM's success in medical image segmentation, SAM 2 presents significant potential for further advancement. However, similar to SAM, SAM 2 is limited by its output of binary masks, inability to infer semantic labels, and dependence on precise prompts for the target object area. Additionally, direct application of SAM and SAM 2 to medical image segmentation tasks yields suboptimal results. In this paper, we explore the upper performance limit of SAM 2 using custom fine-tuning adapters, achieving a Dice Similarity Coefficient (DSC) of 92.30\% on the BTCV dataset~\cite{landman2015miccai}, surpassing the state-of-the-art nnUNet by 12\%. Following this, we address the prompt dependency by investigating various prompt generators. We introduce a UNet to autonomously generate predicted masks and bounding boxes, which serve as input to SAM 2. Subsequent dual-stage refinements by SAM 2 further enhance performance. Extensive experiments show that our method achieves state-of-the-art results on the AMOS2022~\cite{ji2022amos} dataset, with a Dice improvement of 2.9\% compared to nnUNet, and outperforms nnUNet by 6.4\% on the BTCV~\cite{landman2015miccai} dataset.
\end{abstract}

\section{Introduction}
\label{sec:intro}

\begin{figure}[!t]
\centering
\includegraphics[width=0.99\linewidth]{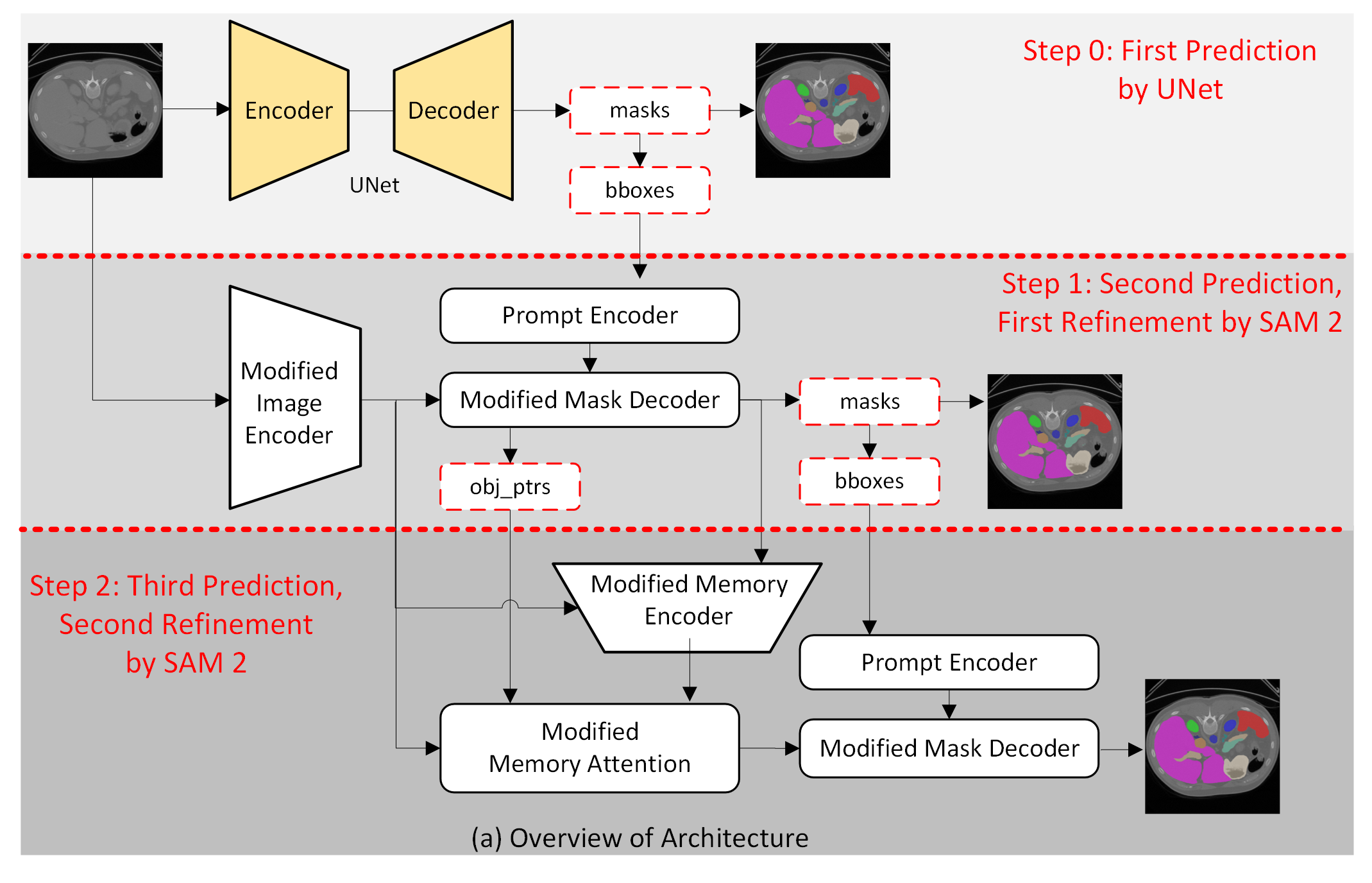}
      \caption{Overview of our proposed RFMedSAM 2.}
\vspace{-0.4cm}
\label{fig:wholeArch}
\end{figure}

Medical image segmentation is vital for biomedical analysis, aiding in disease diagnosis, anomaly detection, and surgical planning. Over recent years, deep learning-based approaches \cite{ronneberger2015u, isensee2019automated, hatamizadeh2022unetr, zhou2021nnformer} have significantly advanced segmentation tasks, with convolutional neural networks (CNNs) and vision transformers (ViTs) becoming the predominant architectures. However, medical imaging datasets often suffer from limited high-quality annotations, which hampers the training of large-scale models. Consequently, architectures with higher inductive biases, such as CNNs, have been more easily trained from scratch to achieve strong performance in medical segmentation tasks.

Foundation models \cite{devlin2018bert, he2022masked}, trained on vast datasets, have shown remarkable capabilities in zero-shot and few-shot generalization across a range of downstream applications \cite{openai2023gpt, radford2021learning}. These models have shifted the paradigm from training task-specific models to a "pre-training then fine-tuning" approach, significantly impacting the field of computer vision. The introduction of the Segment Anything Model (SAM) \cite{kirillov2023segment}, trained on the SA-1B dataset, marked a milestone in prompt-driven natural image segmentation. SAM's success extended to various applications, including medical image segmentation \cite{ma2024segment, xie2024masksam, zhang2024segment, deng2023sam, zhang2023customized, bui2024sam3d}.

Building on this, SAM 2 has been proposed as an enhancement over SAM, extending its functionality to both image and video domains. SAM 2 allows for real-time segmentation across video sequences using a single prompt. Table~\ref{tab:evaluate} shows that SAM 2 performs better than SAM on the BTCV dataset \cite{landman2015miccai}, achieving a Dice score of 82.77\% compared to SAM’s 81.89\%, motivating further exploration into SAM 2 for medical image segmentation tasks.

However, like SAM, SAM 2 has limitations, including its binary mask outputs, the absence of semantic label inference, and reliance on precise prompts for target object identification. Additionally, the performance of SAM and SAM 2 on medical segmentation tasks without modifications falls short of state-of-the-art models.

To address these challenges and maximize SAM 2’s potential for medical image segmentation, we make the following contributions:
\begin{itemize}[leftmargin=*]
\item We introduce RFMedSAM 2, an innovative framework for automatic prompt refinement in medical image segmentation that leverages the multi-stage refinement capabilities of SAM 2.
\item We develop novel adapter modules: a depth-wise convolutional adapter (DWConvAdapter) for attention blocks and a CNN-Adapter for convolutional layers, enhancing spatial information capture and enabling efficient fine-tuning.
\item We establish the upper performance bound of SAM 2 with optimal prompts, achieving a DSC of 92.30\% and surpassing the state-of-the-art nnUNet by 12\% on the BTCV~\cite{landman2015miccai} dataset.
\item We propose an independent UNet for generating masks and bounding boxes as inputs to SAM 2, enabling automatic prompt generation and dual-stage refinement that eliminates the reliance on manual prompts.
\item We perform extensive experiments on challenging medical image datasets (AMOS~\cite{ji2022amos} and BTCV~\cite{landman2015miccai}), demonstrating that RFMedSAM 2 achieves state-of-the-art results, surpassing nnUNet by 2.7\% on the AMOS2022 dataset and 6.4\% on the BTCV dataset.
\end{itemize}

\begin{figure*}[!t]
\centering
  \vspace{-0.6cm}
\includegraphics[width=0.95\textwidth]{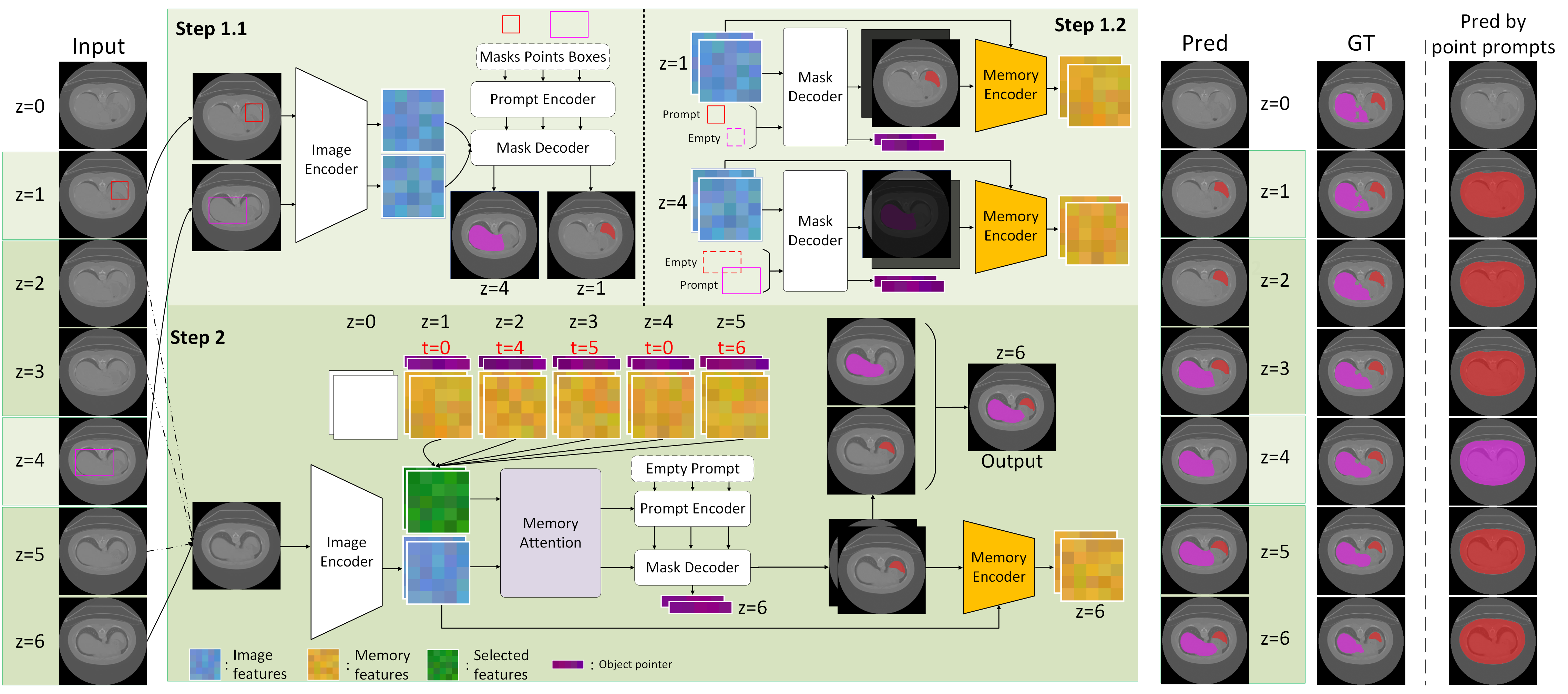}
      \caption{Overview of SAM 2. The pipeline includes steps for processing prompted and unprompted frames.}
\vspace{-0.2cm}
\label{fig:originalSAM2}
\end{figure*}

\section{Related Work}
\label{sec:related}

The field of medical image segmentation has evolved substantially, with traditional machine learning methods giving way to deep learning-based approaches. U-Net \cite{ronneberger2015u} set a new benchmark for medical image segmentation with its encoder-decoder structure and skip connections that help retain spatial context. Following this, nnUNet \cite{isensee2019automated} introduced an automated pipeline that adapts U-Net's architecture to different medical datasets, achieving consistent state-of-the-art results. More recently, transformer-based models, such as UNETR \cite{hatamizadeh2022unetr} and nnFormer \cite{zhou2021nnformer}, have been explored to capture global context and improve accuracy. These models leverage self-attention mechanisms, which help in modeling long-range dependencies, but they often require large datasets for effective training, posing a challenge due to the limited availability of annotated medical images.

Foundation models have transformed the landscape of machine learning by offering a robust starting point for a variety of downstream tasks. The "pre-training then fine-tuning" paradigm has been effective in both natural language processing and computer vision \cite{devlin2018bert, he2022masked}. These models are trained on large, diverse datasets to learn general representations that can be adapted to specific tasks with minimal additional training. This paradigm significantly reduces the reliance on large task-specific datasets and enables zero-shot and few-shot learning. SAM \cite{kirillov2023segment} epitomizes this approach for segmentation tasks by leveraging a pre-trained model that can adapt to new segmentation tasks via prompts. While SAM demonstrated strong zero-shot performance on natural images, its potential in specialized domains like medical imaging sparked interest and subsequent research.

SAM has been extended and tailored for medical image segmentation in several studies. Works such as MedSAM \cite{ma2024segment}, MaskSAM \cite{xie2024masksam}, Self-Prompt SAM~\cite{xie2025selfpromptsammedicalimage}, and other adaptations \cite{zhang2024segment, deng2023sam} highlight the model's flexibility and the community's effort to harness its strengths for medical applications. These adaptations often involve fine-tuning SAM's prompt-encoding mechanisms or integrating domain-specific training strategies to better suit the complexities of medical images, which can include varied resolutions, noise, and non-standardized structures. Despite these advancements, SAM's original design limitations—such as its binary mask outputs and prompt dependency—persist, which restricts its standalone efficacy in comprehensive medical segmentation tasks.

The concept of prompt-driven segmentation introduced by SAM has inspired the development of models that rely on external cues or prompts for segmentation. The approach aligns well with few-shot and zero-shot learning scenarios where annotations are sparse. SAM 2, an extension of SAM, incorporates improvements like memory attention and a memory encoder to process video sequences with greater consistency \cite{kirillov2023segment}. However, these innovations come with challenges, including prompt dependency and limited semantic understanding, which make them less optimal for fully automated medical segmentation tasks. Studies on prompt generation \cite{shaharabany2023autosam} and refinement have shown that integrating mechanisms for automatic prompt generation can reduce the reliance on manually provided prompts and enhance performance in more complex, real-world settings.

The reliance on accurate prompts in SAM 2 and other prompt-driven models presents a clear limitation, particularly in domains where precise annotations are challenging to obtain. Current research is exploring ways to mitigate this dependency, such as designing auxiliary models that can generate reliable prompts or integrating learning mechanisms that adaptively improve prompts during training. Furthermore, memory attention, while effective for maintaining temporal consistency in video segmentation, introduces complexity in terms of training and memory requirements. Addressing these challenges could enable SAM 2 and similar models to reach their full potential in medical image segmentation, bridging the gap between performance and practicality.

\section{The Proposed Approach}
\label{sec:method}

In this section, we first review SAM and SAM 2. Then, we introduce the overall structure of our proposed automatic prompt refinement SAM 2~(RFMedSAM2). Detailed descriptions of each component in RFMedSAM2 can be found in the Appendix.

\subsection{Overview of SAM and SAM 2}
Segment Anything Model (SAM) has proven to be a robust prompt-based foundation model for image segmentation, showcasing strong zero-shot capabilities across various applications. Building on SAM's success, Segment Anything Model 2 (SAM 2) extends these capabilities to both image and video domains, enabling real-time segmentation of objects across entire video sequences using a single prompt.

Both SAM and SAM 2 share a core structure comprising an image encoder, a prompt encoder, and a mask decoder. The image encoder processes input images to generate image embeddings, while the prompt encoder handles input prompts in the form of points, bounding boxes, or masks. The mask decoder then combines image and prompt embeddings to produce binary segmentation masks. SAM employs a Vision Transformer as the backbone of its image encoder, whereas SAM 2 utilizes Hiera~\cite{ryali2023hiera} for enhanced feature representation. SAM 2 also introduces a memory attention module that conditions current frame features on past frames and object pointers, along with a memory encoder that fuses current frame features with output masks to generate memory features.

The SAM 2 pipeline consists of two main stages: the Prompted Frame Processing stage and the Unprompted Frame Processing stage, as illustrated in Figure~\ref{fig:originalSAM2}.
In the prompted frame processing stage, SAM 2 processes frames that contain explicit prompts. Each frame is handled independently, with the prompt guiding the segmentation process. This stage also expands the batch size to match the number of objects expected, ensuring that the output includes masks for each object in the frame. The results from this stage include predicted masks and object pointers, which are passed to the memory encoder to generate memory features.
The unprompted frame processing stage handles frames that do not have explicit prompts. The memory attention module leverages information from previous and prompted frames to build context for segmenting the current frame. In this stage, the prompted frames are assigned a temporal position of 0, while unprompted frames are given temporal positions up to 6, with closer frames having higher temporal positions. This approach helps establish effective context for segmentation, although the original design can struggle with maintaining accurate temporal positioning, potentially leading to errors.

Table~\ref{tab:evaluate} summarizes the performance of SAM and SAM 2 under different settings on the BTCV dataset. The results indicate that SAM 2 outperforms SAM when bounding box prompts are used for each frame, achieving a higher Dice score. This demonstrates the advantage of SAM 2’s enhanced architecture and memory attention capabilities for video segmentation tasks.

\begin{table*}[t]
   \vspace{-0.6cm}
    \centering
    \resizebox{0.99\textwidth}{!}{ 
    \begin{tabular}{@{}cl|c|c|c|c|c|c|c|c|c|c|c|c|c|c@{}}
    \toprule
    & Prompt & \multicolumn{7}{c|}{Bounding boxes as prompts} & \multicolumn{7}{c}{Central points as prompts} \\ \hline
    & Method & \multicolumn{1}{c|}{SAM} & \multicolumn{6}{c|}{SAM 2} & \multicolumn{1}{c|}{SAM} & \multicolumn{6}{c}{SAM 2}  \\ \hline
    & \# frames \slash ~class & \multicolumn{1}{c|}{All} & \multicolumn{2}{c|}{All} & \multicolumn{2}{c|}{Two} & \multicolumn{2}{c|}{One} & \multicolumn{1}{c|}{All} & \multicolumn{2}{c|}{All} & \multicolumn{2}{c|}{Two} & \multicolumn{2}{c}{One}  \\ \hline
    & Frames for Step 2 & -- & All & Unprompted & All & Unprompted& All & Unprompted & -- & All & Unprompted & All & Unprompted & All & Unprompted    \\ \hline
    & DSC~(\%) & 81.89 & 81.17 & 82.77 & 68.75 & 68.03 & 45.00 & 44.07 & 8.86 & 3.81 & 4.90 & 2.11 & 3.43 & 2.53 & 4.59 \\
    \bottomrule
    \end{tabular}}
    \caption{Performance evaluation of SAM and SAM 2 with different prompt settings on the BTCV dataset.}
    \label{tab:evaluate}
   \vspace{-0.2cm}
\end{table*}

\subsection{Analysis and Insights}
SAM 2 offers notable advantages but also has inherent limitations. This section provides a detailed analysis of these aspects. Table~\ref{tab:evaluate} summarizes experiments with various settings for SAM and SAM 2 on the BTCV dataset. All prompts used in these experiments are derived from ground truth, and the models are evaluated without any structural modifications.

1) Bounding box prompts vs. central points: As shown in Table~\ref{tab:evaluate}, the use of central point prompts results in less than 10\% Dice for both SAM and SAM 2. When bounding box prompts are used, the performance significantly improves. For this reason, bounding boxes are used as prompts in subsequent experiments.

2) Per-frame prompts: The results indicate that SAM 2 performs best (82.77\% Dice) when each frame contains a bounding box for every object, underscoring the importance of per-frame prompts for optimal accuracy.

3) Comparison between SAM and SAM 2: With per-frame prompts, SAM achieves a Dice score of 81.89\%, while SAM 2 reaches 82.77\%, demonstrating SAM 2's improved performance over SAM.

4) Step 2 for refinement: Step 2 in SAM 2, which leverages memory attention for unprompted frames, can be extended to all frames for refinement purposes. Forcing Step 2 on all frames results in a slight drop in the Dice score from 82.77\% to 81.17\%, but it shows potential for refining segmentation results. 

\begin{figure*}[!t]
\centering
 \vspace{-0.6cm}
\includegraphics[width=1\textwidth]{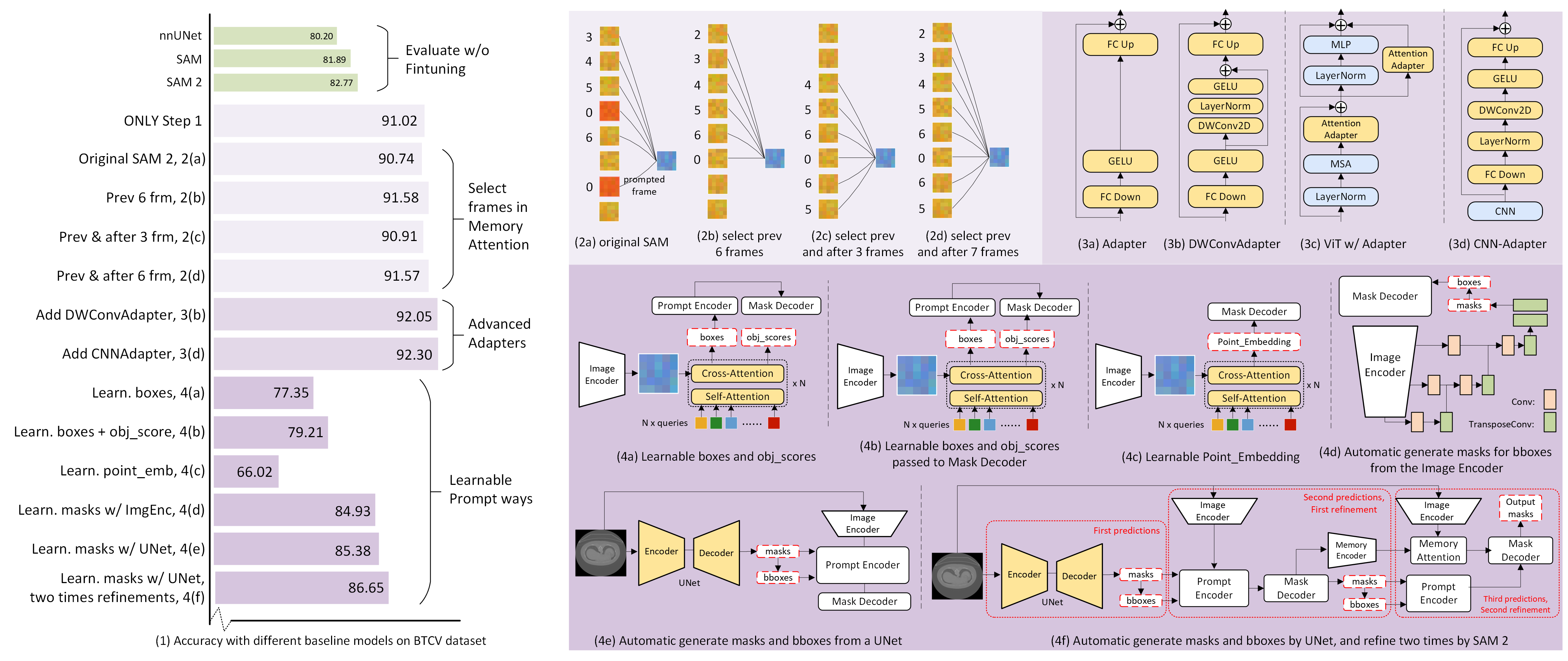}
      \caption{(1) Performance comparisons based on proposed methods. (2) Ablation studies for frame selection strategies. (3) Proposed Adapters. (4) Ablation studies for prompt generators.}
\vspace{-0.4cm}
\label{fig:motivate}
\end{figure*}

5) Streaming operation: Most of SAM 2's modules, except for the memory attention, process images individually without involving temporal operations, which reduces memory usage. The memory attention module stacks features from previous and prompted frames to build connections with the current frame. This method is both efficient and effective, so we maintain this streaming operation in our approach, as shown in Figure~\ref{fig:motivate}.


\subsection{RFMedSAM 2 Architecture}

\subsubsection{Architecture Overview}

Figure~\ref{fig:wholeArch} illustrates the overall architecture of RFMedSAM 2, comprising three primary stages:

\begin{itemize}
\item Initial Prediction Stage: A U-Net model processes the medical images, generating initial multi-class mask predictions. These predictions are converted into bounding boxes to serve as prompts for the next stage.
\item Preliminary Segmentation Stage: The modified image encoder produces image embeddings from the input images, while the prompt encoder converts auxiliary bounding boxes into point embeddings. The mask decoder uses these embeddings to generate initial masks and object pointers. The generated masks are utilized to create new bounding boxes, and the modified memory encoder processes these masks along with current frame features to produce memory features, enabling initial refinement.
\item Refinement Stage: The modified memory attention module takes image features from the encoder and builds relationships with memory features from previous frames. The mask decoder processes these outputs along with new point embeddings from the prompt encoder, producing refined predictions as the final output.
\end{itemize}

\subsubsection{SAM 2 Modifications}

The RFMedSAM 2 design incorporates several key modifications to the SAM 2 architecture:

\begin{itemize}
\item Modified Image Encoder: 
To align various medical imaging modalities with the RGB input format expected by SAM, a sequence of two stacked convolutional layers is added to adapt input modalities. The Hiera~\cite{ryali2023hiera} backbone includes DWConvAdapters in its attention blocks and CNN-Adapters in the FPN module to enhance adaptation.

\item Modified Mask Decoder:
The mask decoder includes adapters after self- and cross-attention blocks and in parallel with MLP layers to capture spatial information more effectively. DWConvAdapters facilitate spatial learning, while CNN-Adapters adapt convolutional layers for medical image processing.

\item UNet, Memory Encoder, and Memory Attention:
The U-Net maintains a symmetric encoder-decoder structure with skip connections for better spatial detail retention. The memory encoder integrates CNN-Adapters to adapt its components for processing medical image features. The memory attention module incorporates DWConvAdapters within its transformer blocks to process spatial information effectively.
\end{itemize}

%

\subsection{Architectural Design}

In this section, we present our architectural advancements for SAM 2 aimed at enhancing its performance in medical image segmentation. Our main focus is on designing an improved memory attention strategy and novel adapters to maximize SAM 2's segmentation capabilities. While ground-truth (GT) prompts are used to explore the upper bound of performance, the key contributions lie in the architectural modifications that support robust fine-tuning and improved adaptability.

\subsubsection{Refined Frame Selection Strategy}
Step 2 in SAM 2, originally used for processing unprompted frames in memory attention, is a crucial refinement step for enhancing segmentation consistency. Our goal was to extend this step to process all frames and refine predictions across the entire sequence. In the original design, SAM 2 assigns a temporal position of 0 to all prompted frames, leading to ambiguous temporal positioning and the potential for false positives when attention is applied across frames.

To optimize the memory attention strategy, we experimented with different frame selection methods and temporal position assignments using a baseline model. Figure~\ref{fig:motivate}(2) illustrates four strategies. The original strategy (Figure~\ref{fig:motivate}(2a)) achieved a Dice Similarity Coefficient (DSC) of 90.74\%, which was outperformed by a simpler approach using only Step 1, indicating limitations due to incorrect temporal positioning.

Our improved strategy, shown in Figure~\ref{fig:motivate}(2b), assigns the current frame index as the temporal position of 0, ensuring that the model prioritizes memory features of the current frame that include Step 1 mask predictions. This strategy significantly enhances performance, achieving a DSC of 91.58\%. Alternative strategies that select both forward and backward frames (Figures~\ref{fig:motivate}(2c) and \ref{fig:motivate}(2d)) resulted in either performance drops or increased memory requirements.

We adopt the frame selection method from Figure~\ref{fig:motivate}(2b), selecting up to 6 previous frames and setting the current frame index as the temporal position of 0. This approach ensures comprehensive memory integration for robust segmentation refinement.

\subsubsection{Design of Novel Adapters for Enhanced Fine-Tuning}
To enable parameter-efficient fine-tuning while retaining SAM 2's zero-shot capabilities, we designed new adaptation mechanisms that enhance spatial and convolutional processing within SAM 2’s architecture:
\begin{itemize}
\item Depth-wise Convolutional Adapter (DWConvAdapter): The image encoder, memory attention, and mask decoder contain attention blocks that process image embeddings with rich spatial information. To strengthen this, we introduce the DWConvAdapter (Figure~\ref{fig:motivate}(3b)), which incorporates depth-wise convolutions to capture spatial context effectively. Integrating DWConvAdapters improved the DSC by 0.47\%, demonstrating its utility in enhancing spatial learning.

\item CNN-Adapter for Convolutional Layers: Given the presence of multiple convolutional layers in SAM 2, we also developed a CNN-Adapter to facilitate better adaptation within these layers (Figure~\ref{fig:motivate}(3b)). The addition of CNN-Adapters led to a DSC increase of 0.25\%, further validating the effectiveness of targeted architectural modifications.
\end{itemize}

Our final model incorporates original adapters for point embedding attention blocks, DWConvAdapters for image embedding attention blocks, and CNN-Adapters for convolutional layers. This comprehensive architecture allows SAM 2 to achieve over a 4\% improvement compared to state-of-the-art methods, as shown in Table~\ref{tab:amos}, establishing its capability for advanced medical image segmentation.

\begin{table*}[!t]\small
    \setlength{\tabcolsep}{3pt}
    \centering
    \vspace{-0.6cm}
    \resizebox{1\linewidth}{!}{ 
    \begin{tabular}{@{}c|c|l|ccccccccccccccc|c@{}}
    \toprule
    Semantic labels & Prompts & Method & Spl. & R.Kd & L.Kd & GB & Eso. & Liver & Stom. & Aorta & IVC  & Panc. & RAG & LAG & Duo. & Blad. &  Pros. & Average \\
    \midrule
    & & TransBTS \cite{wang2021transbts} & 0.885 & 0.931 & 0.916 & 0.817 & 0.744 & 0.969 & 0.837 & 0.914 & 0.855 & 0.724 & 0.630 & 0.566 & 0.704 & 0.741 & 0.650 & 0.792 \\
    & & UNETR \cite{hatamizadeh2022unetr} & 0.926 & 0.936 & 0.918 & 0.785 & 0.702 & 0.969 & 0.788 & 0.893 & 0.828 & 0.732 & 0.717 & 0.554 & 0.658 & 0.683 & 0.722 & 0.762 \\
    \CheckmarkBold & -- & nnFormer \cite{zhou2021nnformer} & 0.935 & 0.904 & 0.887 & 0.836 & 0.712 & 0.964 & 0.798 & 0.901 & 0.821 & 0.734 & 0.665 & 0.587 & 0.641 & 0.744 & 0.714 & 0.790 \\
    & & SwinUNETR \cite{hatamizadeh2021swin} & 0.959 & 0.960 & 0.949 & \textbf{0.894} & 0.827 & 0.979 & 0.899 & 0.944 & 0.899 & 0.828 & \textbf{0.791} & 0.745 & 0.817 & 0.875 & 0.841 & 0.880 \\
    & & nn-UNet~\cite{isensee2019automated} & 0.965 & 0.959 & 0.951 & 0.889 & 0.820 & \textbf{0.980} & 0.890 & 0.948 & 0.901 & 0.821 & 0.785 & 0.739 & 0.806 & 0.869 & 0.839 & 0.878 \\
    \hline
    \XSolidBrush & nnUNet & SAM~\cite{kirillov2023segment} bbox & 0.679 & 0.741 & 0.640 & 0.168 & 0.443 & 0.773 & 0.671 & 0.651 & 0.554 & 0.434 & 0.232 & 0.324 & 0.444 & 0.698 & 0.602 & 0.538 \\
    \XSolidBrush & nnUNet & SAM 2~\cite{ravi2024sam} bbox & 0.784 & 0.817 & 0.819 & 0.664 & 0.734 & 0.780 & 0.697 & 0.793 & 0.739 & 0.536 & 0.457 & 0.604 & 0.563 & 0.744 & 0.691 & 0.695 \\
    \XSolidBrush & nnUNet & MedSAM~\cite{ma2024segment} bbox & 0.714 & 0.811 & 0.702 & 0.193 & 0.469 & 0.759 & 0.725 & 0.701 & 0.681 & 0.434 & 0.365 & 0.412 & 0.462 & 0.783 & 0.758 & 0.600 \\
    \CheckmarkBold & No needs & SAMed~\cite{zhang2023customized} & 0.849 & 0.857 & 0.830 & 0.573 & 0.733 & 0.894 & 0.816 & 0.855 & 0.784 & 0.727 & 0.622 & 0.683 & 0.701 & 0.844 & 0.819 & 0.772 \\
    \CheckmarkBold & No needs & SAM3D~\cite{bui2024sam3d} & 0.796 & 0.863 & 0.871 & 0.428 & 0.711 & 0.908 & 0.833 & 0.878 & 0.749 & 0.699 & 0.564 & 0.607 & 0.635 & 0.884 & 0.840 & 0.751 \\
    \hline
    \hline
    \CheckmarkBold & No needs & RFMedSAM 2   & \textbf{0.972} & \textbf{0.971} & \textbf{0.966} & 0.887 & \textbf{0.878} & \textbf{0.980} & \textbf{0.943} & \textbf{0.958} & \textbf{0.925} & \textbf{0.896} & 0.781 & \textbf{0.811} & \textbf{0.853} & \textbf{0.921} & \textbf{0.859} & \textbf{0.907} \\
    \bottomrule
    \end{tabular}
    }
    \caption{Comparison of RFMedSAM 2 with SOTA methods on the AMOS testing dataset, evaluated using Dice Score. All results are based on 5-fold cross-validation without ensemble techniques. ``Semantic labels'' indicate the model's ability to infer semantic labels, while ``Prompts'' specify the prompt source. The best results are shown in \textbf{bold}.}
    \label{tab:amos}
\vspace{-0.4cm}
\end{table*}

\subsection{Advancing Prompt Generation}

After exploring the upper performance limit of SAM 2 with accurate ground truth (GT) prompts, the next step was to develop a practical solution that removes reliance on such precise prompts, which are unrealistic for real-world medical image segmentation. Recognizing that SAM 2 can achieve exceptional performance with accurate prompts, we proposed a prompt generation framework that refines both generated prompts and final predictions during training. Six distinct blocks for automatic prompt generation were designed, as shown in Figure~\ref{fig:motivate}(4a)-(4f), categorized into two main types: learnable point coordinate representations (Figures~\ref{fig:motivate}(4a)-(4c)) and learnable masks (Figures~\ref{fig:motivate}(4d)-(4f)). Performance results for these blocks are shown in the last six bars of Figure~\ref{fig:motivate}(1).

\subsubsection{Learnable Point Coordinate Representations} 
The block depicted in Figure~\ref{fig:motivate}(4a) initializes object queries for each class, which are processed through a series of self-attention and cross-attention blocks that interact with the current image features. Multiple MLP layers are used to adjust the embedding dimensions for generating box coordinates and object scores. SAM 2 employs stricter label criteria for point prompts than its predecessor, using labels such as no object~(-1), negative/positive points~(0, 1), and box prompts~(2, 3). In prior experiments, GT prompts included labels to denote the absence of an object at specific frames. In the current approach, object scores are trained to indicate whether a given frame should contain a prompt or none at all.

Despite these efforts, the block in Figure~\ref{fig:motivate}(4a) only reached a DSC of 77.35\%, indicating a significant performance gap. Integrating object scores from the mask decoder, as seen in Figure~\ref{fig:motivate}(4b), increased performance by 1.9\%, but the results remained below expectations. To bypass the challenges of label representation, we designed a learnable point embedding block, allowing it to learn coordinate and label representations directly (Figure~\ref{fig:motivate}(4c)). However, this approach resulted in an 11\% drop in DSC, highlighting the difficulty of learning precise prompts.

Accurate coordinate prediction, essential for bounding box prompts, proved challenging due to the non-coordinate-encoded nature of image embeddings and the random initialization of embeddings. Additionally, bounding boxes lacked the semantic richness necessary for effective multi-class segmentation. This led us to pivot towards using learnable masks, which offer more robust semantic information.

\subsubsection{Learnable Masks} 
We found that predicting masks first and deriving bounding boxes from them provided a more reliable approach than directly predicting coordinates. The structure illustrated in Figure~\ref{fig:motivate}(4d) incorporates a hierarchical design of convolutional layers combined with multi-level features from the image encoder. Starting with lower-resolution features, the model progressively increases resolution through convolutional layers, combining them with higher-resolution features. Auxiliary loss functions supervise the generated masks by comparing them to the ground truth, achieving a DSC of 84.93\%. Although this was an improvement, it fell short of top-tier performance.

One challenge was that both auxiliary losses from the generated masks and final output losses from SAM 2 impacted updates to the image encoder, leading to conflicts that hindered optimal training. The distinct architectures between the prompt generator and SAM 2 complicated synchronized parameter updates, making it difficult to maintain balance and achieve consistent improvements.

To overcome this, we introduced an independent U-Net architecture alongside SAM 2 to generate masks that do not interfere with SAM 2’s parameter updates (Figure~\ref{fig:motivate}(4e)). This U-Net-generated mask was used to derive bounding boxes as input prompts for SAM 2, raising the performance to 85.38\%. To further enhance the interaction between the U-Net and SAM 2, we routed the masks and bounding boxes directly into the first step of SAM 2, enabling the prediction of a refined set of masks and updated bounding boxes. These were subsequently fed into the second step for further refinement, resulting in an overall performance of 86.48\% DSC.

This multi-stage prompt generation and refinement pipeline significantly reduces reliance on precise GT prompts and emphasizes the model's capacity for self-sufficient prompt generation in realistic medical imaging scenarios.




\section{Experimental Evaluation}
\label{sec:experiments}

\begin{table*}[!t]\small
    \vspace{-0.6cm}
    \setlength{\tabcolsep}{3pt}
    \centering
    \resizebox{0.9\linewidth}{!}{ 
    \begin{tabular}{@{}c|c|l|cccccccccccc|c@{}}
    \toprule
    Semantic labels & Prompts & Method & Spl. & R.Kd & L.Kd & GB & Eso. & Liv. & Stom. & Aorta & IVC & Veins & Panc. & AG & DSC \\
    \midrule

     & & TransUNet~\cite{chen2021transunet}  & 0.952  & 0.927 & 0.929 & 0.662 & 0.757 & 0.969 & 0.889 & 0.920 & 0.833 & 0.791 & 0.775 & 0.637 & 0.838 \\ 
    & & 3D UX-Net~\cite{lee20223d}  & 0.946 & 0.942 & 0.943 & 0.593 & 0.722 & 0.964 & 0.734 & 0.872 & 0.849 & 0.722 & 0.809 & 0.671 & 0.814 \\
    & & UNETR~\cite{hatamizadeh2022unetr} & 0.968 & 0.924 & 0.941 & 0.750 & 0.766 & 0.971 & 0.913 & 0.890 & 0.847 & 0.788 & 0.767 & 0.741 & 0.856 \\ 
    \CheckmarkBold & -- & Swin-UNETR~\cite{hatamizadeh2021swin} & 0.971 & 0.936 & 0.943 & 0.794 & 0.773 & 0.975 & 0.921 & 0.892 & 0.853 & 0.812 & 0.794 & 0.765 & 0.869 \\
    & & nnUNet~\cite{isensee2019automated} & 0.942 & 0.894 & 0.910 & 0.704 & 0.723 & 0.948 & 0.824 & 0.877 & 0.782 & 0.720 & 0.680 & 0.616 & 0.802 \\
    & & nnFormer~\cite{zhou2021nnformer}  & 0.935 & 0.949 & 0.950 & 0.641 & 0.795 & 0.968 & 0.901 & 0.897 & 0.859 & 0.778 & 0.856 & 0.739  & 0.856 \\ 
    \hline

    \XSolidBrush & GT & SAM~\cite{kirillov2023segment} & 0.933 & 0.922 & 0.927 & 0.805 & 0.831 & 0.899 & 0.808 & 0.890 & 0.894 & 0.492 & 0.728 & 0.708 & 0.819 \\
    \XSolidBrush & GT & SAM 2~\cite{ravi2024sam} & 0.946 & 0.923 & 0.924 & 0.859 & 0.888 & 0.928 & 0.893 & 0.852 & 0.884 & 0.434 & 0.694 & 0.705 & 0.828 \\
    \XSolidBrush & GT & MedSAM~\cite{ma2024segment} & 0.751 & 0.814 & 0.885 & 0.766 & 0.721 & 0.901 & 0.855 & 0.872 & 0.746 & 0.771 & 0.760 & 0.705 & 0.803 \\
    
    \XSolidBrush & GT & SAM-U~\cite{deng2023sam} & 0.868 & 0.776 & 0.834 & 0.690 & 0.710 & 0.922 & 0.805 & 0.863 & 0.844 & 0.782 & 0.611 & 0.780 & 0.790 \\
    \XSolidBrush & GT & SAM-Med2D~\cite{cheng2023sam} & 0.873 & 0.884 & 0.932 & 0.795 & 0.790 & 0.943 & 0.889 & 0.872 & 0.796 & 0.813 & 0.779 & 0.797 & 0.847 \\

     \XSolidBrush & GT & RFMedSAM 2 & 0.961 & 0.943 & 0.945 & 0.909 & 0.918 & 0.965 & 0.945 & 0.954 & 0.942 & 0.968 & 0.883 & 0.843 & 0.923 \\ \hline
    
    \CheckmarkBold & No Needs &SAMed~\cite{zhang2023customized} & 0.862 & 0.710 & 0.798 & 0.677 & 0.735 & 0.944 & 0.766 & 0.874 & 0.798 & 0.775 & 0.579 & 0.790 & 0.776 \\

    \CheckmarkBold & No Needs & SAM3D~\cite{bui2024sam3d} & 0.933 & 0.901 & 0.909 & 0.601 & 0.733 & 0.944 & 0.882 & 0.856 & 0.778 & 0.722 & 0.759 & 0.590 & 0.801 \\

    \CheckmarkBold & No Needs & RFMedSAM 2 & 0.969 & 0.947 & 0.953 & 0.611 & 0.817 & 0.974 & 0.909 & 0.917 & 0.887 & 0.803 & 0.865 & 0.747 & 0.867 \\ 
    \bottomrule
    \end{tabular}
    }
    \caption{
    Comparison of RFMedSAM 2 with state-of-the-art methods on the BTCV dataset. ``Semantic labels'' indicate the model's ability to infer labels, while ``Prompt'' specifies the source of the prompt.}
    \vspace{-0.2cm}
    \label{tab:sotaSynapse}
\end{table*}

\begin{figure*}[!t]
\centering
\includegraphics[width=1\textwidth]{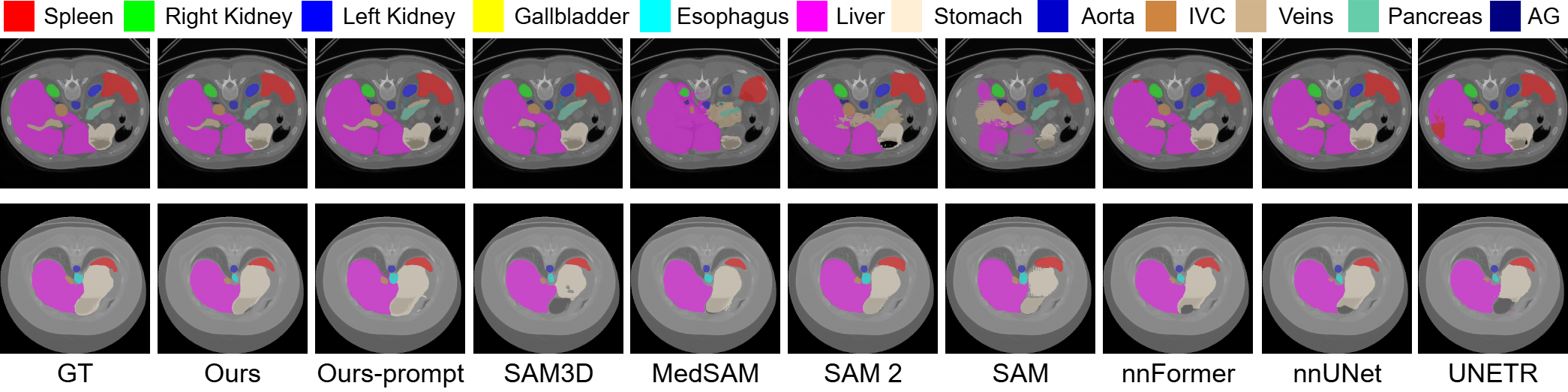}
      \caption{Qualitative comparison on BTCV dataset. RFMedSAM 2 is the most precise for each class and has fewer segmentation outliers.} 
\vspace{-0.4cm}
\label{fig:visualizationBTCV}
\end{figure*}

\subsection{Datasets and Evaluation Metrics.}

We conduct experiments using two publicly available datasets: the AMOS22 Abdominal CT Organ Segmentation dataset~\cite{ji2022amos} and the Beyond the Cranial Vault (BTCV) challenge dataset~\cite{landman2015miccai}. \textbf{(i)} The AMOS22 dataset contains 200 abdominal CT scans with manual annotations for 16 anatomical structures, which serve as the basis for multi-organ segmentation tasks. The testing set comprises 200 images, and we evaluate our model using the AMOS22 leaderboard. \textbf{(ii)} The BTCV dataset includes 30 cases of abdominal CT scans. Following established split strategies~\cite{hatamizadeh2021swin}, we use 24 cases for training and 4 cases for validation. Performance is assessed using the average Dice Similarity Coefficient (DSC) across 13 abdominal organs.

In Tables~\ref{tab:amos} and~\ref{tab:sotaSynapse}, ``Semantic labels'' refer to the ability of a model to infer and predict labels, while ``Prompt'' specifies the prompt source. Since SAM and MedSAM do not predict semantic labels and require additional prompts, we use GT or predictions inferred by a pre-trained nnUNet to generate prompts, with the corresponding labels used as semantic labels.

\subsection{Comparison with State-of-the-Art Methods}

\subsubsection{Results on the AMOS22 Dataset.} 
Table~\ref{tab:amos} presents the quantitative results on the AMOS22 dataset, comparing our proposed RFMedSAM 2 with widely recognized segmentation methods, including CNN-based methods (nnUNet~\cite{isensee2019automated}), transformer-based methods (UNETR~\cite{hatamizadeh2022unetr}, SwinUNETR~\cite{hatamizadeh2021swin}, nnFormer~\cite{zhou2021nnformer}), and SAM-based methods (SAM~\cite{kirillov2023segment}, SAM 2~\cite{ravi2024sam}, MedSAM~\cite{ma2024segment}, SAMed~\cite{zhang2023customized}, and SAM3D~\cite{bui2024sam3d}). To ensure fairness, all methods are evaluated using 5-fold cross-validation without ensemble techniques.

We observe that our RFMedSAM 2 outperforms all existing methods on most organs, achieving a new state-of-the-art performance in DSC. When utilizing the predictions from nnUNet for bounding box prompts, SAM, SAM 2, and MedSAM exhibit decreases of 34\%, 18\%, and 27\%, respectively, compared to nnUNet's accuracy of 87.8\%. These reductions in accuracy indicate negative implications for the results. SAM 2 achieves the best performance, which demonstrates it presents the strongest zero-shot capabilities. Specifically, RFMedSAM 2 surpasses nnUNet by 2.9\% in DSC, respectively. RFMedSAM 2 surpasses SAMed and SAM3D by 23\% and 25\% in DSC, respectively. The significant improvement demonstrates our proposed prompt-free RFMedSAM 2 is better than other prompt-free SAM models.
In the extremely hard AMOS 2022 dataset, our RFMedSAM 2 achieves state-of-the-art performance, which confirms the efficacy of our method. 

\subsubsection{Results on the BTCV Dataset.} 
Table~\ref{tab:sotaSynapse} shows the quantitative performance on the BTCV dataset, comparing RFMedSAM 2 with leading SAM-based methods with proper prompts(\textit{i.e.},~SAM~\cite{kirillov2023segment}, SAM2~\cite{ravi2024sam}, MedSAM~\cite{ma2024segment}, SAM-U~\cite{deng2023sam}, and SAM-Med2D~\cite{cheng2023sam}), SAM-based methods without prompts (\textit{i.e.},~SAMed~\cite{zhang2023customized} and SAM3D~\cite{bui2023sam3d}), convolution-based methods (VNet~\cite{ronneberger2015u} and nnUNet~\cite{isensee2019automated}), transformer-based methods (TransUNet~\cite{chen2021transunet}, SwinUNet~\cite{cao2021swin}, 
and nnFormer~\cite{zhou2021nnformer}). 
We observe that RFMedSAM 2 outperforms all existing methods, setting a new state-of-the-art benchmark. When provided with proper prompts, RFMedSAM 2 achieves a DSC of 92.3\%, representing a significant 5\% improvement over the previous state-of-the-art method. In comparison, among SAM-based methods with the proper prompts, the best performance, achieved by SAM-Med2D, reaches only 84.7\%. Our proposed RFMedSAM 2 surpasses this by 7.6\%, highlighting its superior effectiveness over SAM-based methods with prompts. When prompts are not provided, our proposed prompt-free RFMedSAM 2 outperforms the other prompt-free SAMed and SAM3D by 9\% and 6\%, respectively. Compared with non-SAM-based methods, our method surpasses nnUNet and nnFormer by 6.4\% and 1\% in DSC for the highly saturated dataset. In Figure~\ref{fig:visualizationBTCV}, we illustrate qualitative results compared to representative methods. These results also demonstrate that our RFMedSAM 2 can predict more accurately the `Stomach', `Spleen', and `Liver' labels. 

\begin{table}
    \vspace{-0.6cm}
    \centering
    \resizebox{0.92\linewidth}{!}{ 
    \begin{tabular}{@{}c|c|c|c}
    \toprule
     & train with prompts & learnable bboxes  & learnable masks  \\ \midrule
    
    w/ obj\_score & 0.923 & 0.792 & 0.847  \\ \hline
    w/o obj\_score & 0.920 & 0.628 & 0.867 \\
    \bottomrule
    \end{tabular}} 
    \caption{Experiments for different models with and without the prediction of object scores on BTCV dataset.}
    \label{tab:analysis_obj_score}
\end{table}

\begin{table}
    \vspace{-0.2cm}
    \centering
    \resizebox{0.72\linewidth}{!}{ 
    \begin{tabular}{@{}c|c|c|c}
    \toprule
    Dataset & Step 0 - UNet & Step 1 - SAM  & Step 2 - SAM  \\ \midrule
    
    BTCV & 0.856 & 0.864 & 0.867  \\ \hline
    AMOS & 0.895 & 0.898 & 0.907 \\
    \bottomrule
    \end{tabular}} 
    \caption{The performance of output predictions for different steps. Two refinements lead to gradual improvement.}
    \label{tab:analysis_refinement}
    \vspace{-0.1cm}
\end{table}

\begin{figure}[!t]
\centering
  \vspace{-0.1cm}
\includegraphics[width=0.99\linewidth]{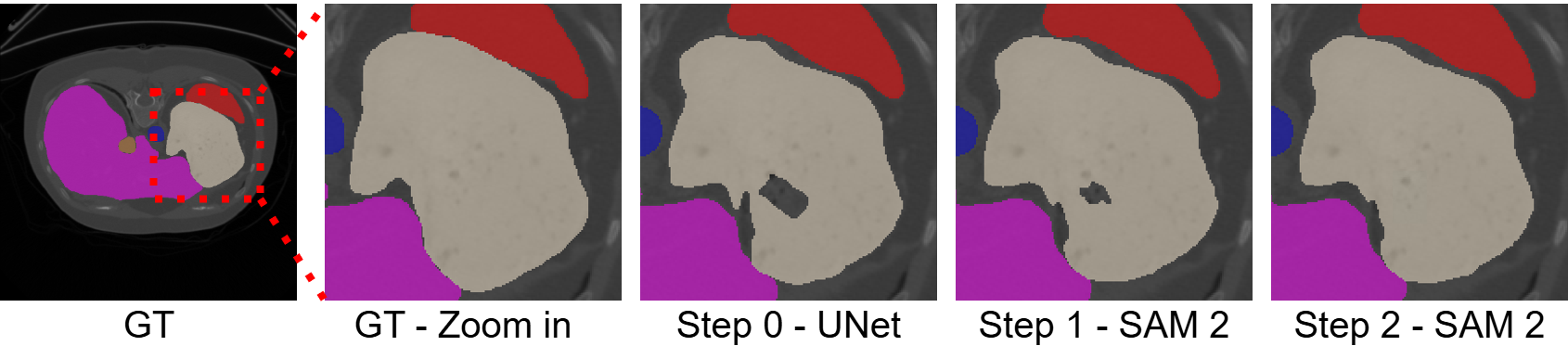}
      \caption{Comparisons with different output predictions for Step 0, Step 1, and Step 2.} 
\vspace{-0.6cm}
\label{fig:refinement_comparison}
\end{figure}

\subsection{Analysis}
\noindent \textbf{Refinements.}
Table~\ref{tab:analysis_refinement} presents experimental results for output predictions at different steps on the BTCV and AMOS datasets. The results show a gradual improvement in performance, starting from the initial prediction at Step 0~(UNet), followed by the second prediction after the first refinement at Step 1~(SAM 2), and finally the third prediction after the second refinement at Step 2~(SAM 2). 
Figure~\ref{fig:refinement_comparison} visualizes these comparisons across the three different steps. The results clearly demonstrate how the hole is progressively filled through the two refinements, highlighting the effectiveness of our model's refinement process.

\noindent \textbf{Object score.} 
We experiment with three different baseline models: fine-tuning of SAM 2 with prompts, with learnable bounding boxes as the prompt generator, and with learnable masks as the prompt generator, both with and without the prediction of object scores. Table~\ref{tab:analysis_obj_score} shows that i) learning object scores with prompts does not significantly improve performance compared to using prompts without object scores, as the prompt itself indicates whether the object exists in a given frame. ii) The model with learnable bounding boxes benefits from learning object scores, as the accuracy of bounding box predictions is generally not high. iii) The model with learnable masks shows worse performance when learning object scores, as the probability distribution of the output predicted masks provides more accurate predictions. The object scores, which directly determine a single probability plane, can negatively impact this accuracy.

\begin{table}
    \vspace{-0.6cm}
    \centering
    \resizebox{0.68\linewidth}{!}{ 
    \begin{tabular}{@{}c|c|c|c}
    \toprule
     & (2, 1024, 1024) & (8, 512, 512)  & (32, 256, 256)  \\ \midrule
    
    DSC & 0.751 & 0.827 & 0.867  \\ 
    \bottomrule
    \end{tabular}} 
    \caption{Experiments for different input patch sizes on BTCV.}
    \label{tab:analysis_patch_size}
\end{table}

\begin{table}
    \vspace{-0.2cm}
    \centering
    \resizebox{0.95\linewidth}{!}{ 
    \begin{tabular}{@{}c|c|c|c|c}
    \toprule
     & 3D UNet & 2D UNet  & 2D UNet + Attention & 3D UNet + Attention  \\ \midrule
    
    DSC & 0.825 & 0.807 & 0.805 & 0.815  \\ 
    \bottomrule
    \end{tabular}} 
    \caption{Experiments for different UNet models on BTCV dataset.}
    \label{tab:analysis_UNet}
    \vspace{-0.6cm}
\end{table}

\noindent \textbf{Input Patch Sizes and UNet Choices.} Table~\ref{tab:analysis_patch_size} shows the performance of different input patch sizes with the same number of pixels. Increasing the number of depth dimensions can bring benefits. Table~\ref{tab:analysis_UNet} illustrates different U-Net architectures. It shows 3D UNet is better than 2D UNet since the depth dimension can be learned. Involving attention blocks in the bottleneck can not bring benefits due to a strong inductive bias for medical image segmentation.

\section{Conclusion}
\label{sec:conclusion}

In this paper, we introduced RFMedSAM 2, a novel framework for automatic prompt refinement that extends the SAM 2 pipeline to facilitate multiple refinement stages, enabling its adaptation for volumetric medical image segmentation. We explored two main branches to fully leverage SAM 2's potential. 

The first branch focused on evaluating the upper performance bound of SAM 2 when provided with accurate prompts. To enhance the model, we proposed depth-wise convolutional adapters (DWConvAdapters) for attention blocks involving image embeddings to capture spatial information, as well as CNN-Adapters for convolutional layers to enable efficient fine-tuning. With these adapters and optimized memory attention positioning, our model achieved a Dice Similarity Coefficient (DSC) of 92.30\%, surpassing the nnUNet by 12\% on the BTCV~\cite{landman2015miccai} dataset.

The second branch aimed to overcome the reliance on precise prompts by designing a module capable of generating accurate prompts automatically. Building on the insights gained from determining the upper bound and modifying SAM 2, we proposed an independent U-Net to predict masks and bounding boxes, which serve as input prompts for SAM 2. These prompts underwent two refinement stages within SAM 2, further enhancing performance. Our model achieves a DSC of 90.7\% and 86.7\% on the AMOS2022~\cite{ji2022amos} and BTCV~\cite{landman2015miccai} dataset, respectively.

Overall, our comprehensive approach demonstrates the effectiveness of RFMedSAM 2 in achieving state-of-the-art results in medical image segmentation.
For future work, we plan to explore the extension of RFMedSAM 2 to other types of medical imaging modalities, such as MRI and ultrasound, and investigate its potential for real-time clinical applications. 


{
    \small
    \bibliographystyle{ieeenat_fullname}
    \bibliography{main}
}

\clearpage
\setcounter{page}{1}
\maketitlesupplementary

\section{Intrinsic issues of SAM2}
Figure~\ref{fig:originalSAM2} illustrates the whole pipeline of SAM 2, highlighting several intrinsic issues for medical image segmentation. 

i) \textbf{Omission to predict the first few frames:} The first frame in two objects is the second frame, therefore, SAM 2 begins processing from the second frame, disregarding the first frame, even though it contains both objects.

ii) \textbf{Empty prompt affecting object prediction:} When no prompt is provided for an object, but the object is still present, the empty prompt restricts prediction for that object. For instance, at frames $z=1$ and $z=4$, the purple and red objects, respectively, are omitted from the predictions. 

iii) \textbf{Confusion of temporal positions:} All prompted frames are assigned a temporal position of 0. While this approach increases attention to the prompted frames, it loses the relative temporal positioning of all the prompt frames. Moreover, since SAM 2 skips over prompted frames, the relative temporal positions of the unprompted frames are distorted. 
For example, the real relative temporal position of the frame $z=3$ with respect to the current frame $z=6$ should be 3, but due to the prompted frame at $z=4$, the relative temporal position is incorrectly assigned as 2.

\section{Potentials to force Step 2 for all frames.}
 When we provide prompts at each frame for each class, SAM 2 does not process Step 2 and does not leverage the capabilities of Memory Attention, which can build relations with previous frames and prompted frames. To explore this functionality, we force Step 2 for all frames after processing Step 1.  
\begin{figure}[!hbtp]
\centering
  \vspace{-0.2cm}
\includegraphics[width=1\linewidth]{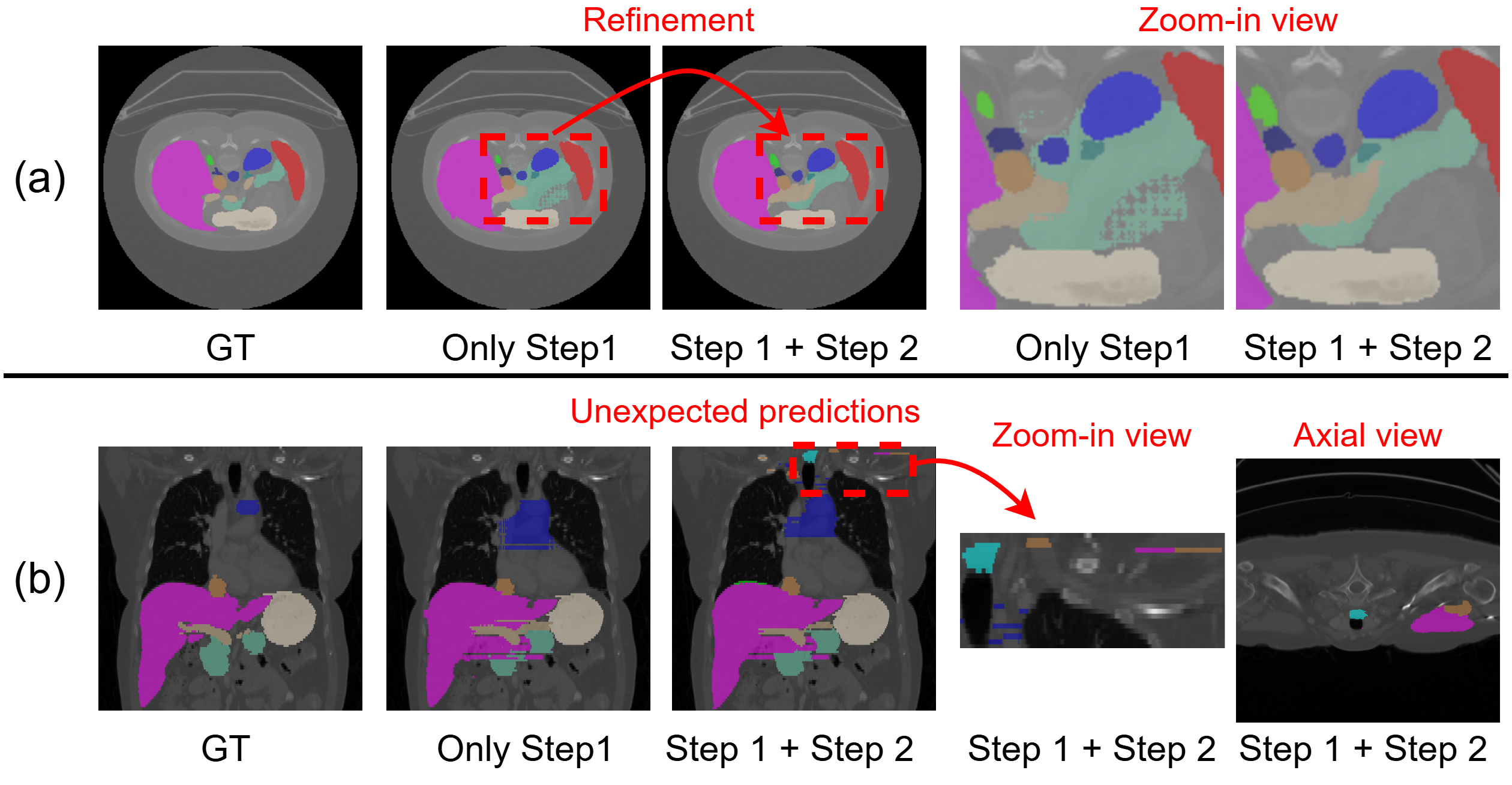}
      \caption{Benefits for refinement by Step 2.} 
\vspace{-0.2cm}
\label{fig:refinement}
\end{figure}
Although the results decreased slightly from 82.77\% to 81.17\% Dice, we find a potential refinement benefit illustrated in Figure~\ref{fig:refinement}(a). Through Step 2, the green area is refined and becomes more accurate, demonstrating the significant potential of the refinement process. As a result, we plan to incorporate this approach into our method. However, the refinement introduced by Step 2 also has some drawbacks. In Figure~\ref{fig:refinement}(b), we show orthogonal planes in relation to the axial plane~(a sequence of the axial plane images is fed to SAM 2). The top portion of Figure~\ref{fig:refinement}(b) presents unexpected predictions. 
Since SAM 2 assigns a temporal position of 0 to the prompted frames, which are always involved in memory attention, the incorrect relative temporal positioning leads to these unexpected and incorrect predictions. We will address this issue in the next section.

\begin{figure*}[!t]
\centering
\includegraphics[width=1\linewidth]{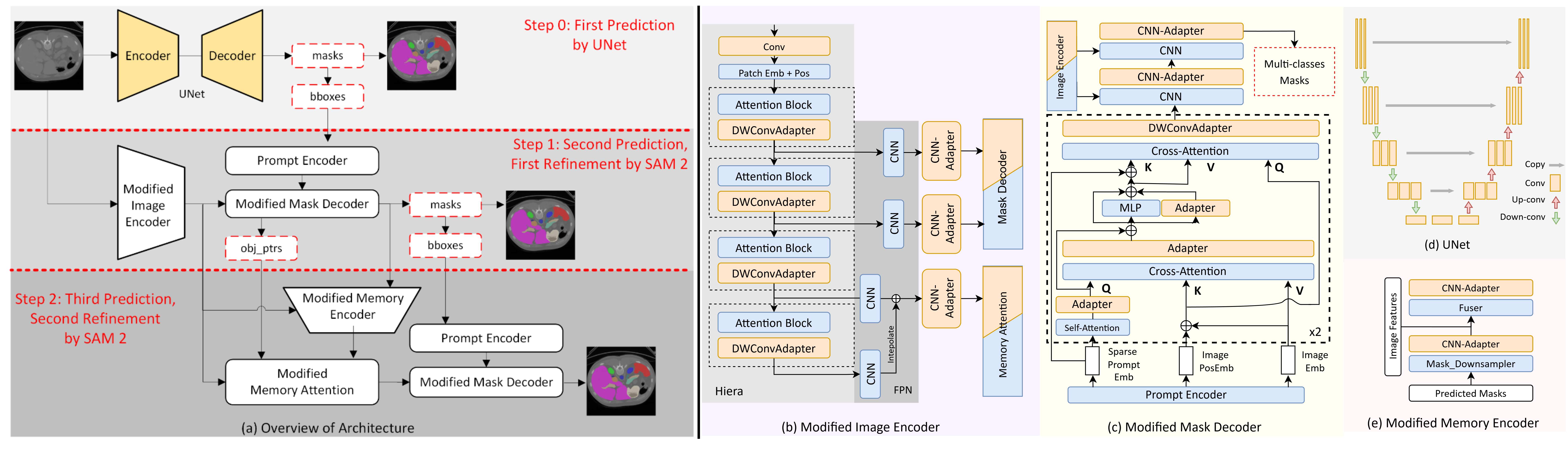}
      \caption{Details of the whole architecture of RFMedSAM 2.} 
\vspace{-0.6cm}
\label{fig:wholeArch_details}
\end{figure*}

\section{Motivation Behind the Designed Adapters.} 
Since the image encoder, the memory attention, and the mask decoder contain attention blocks for image embedding, which includes significant spatial information. Therefore, we design the depth-wise convolutional adaption~(DWConvAdapter) illustrated in Figure~\ref{fig:motivate}(3b) to learn spatial information. After using DWConvAdapters for the attention blocks with image embedding, the performance increases by 0.47\%. The motivation behind the DWConvAdapter design is to extend the original adapter by incorporating a depth-wise convolution layer, followed by layer normalization and a GeLU activation function, to effectively learn spatial information. A parallel skip connection is included to preserve the original structure. In the worst case, where the depth-wise layer learns nothing~(\textit{i.e.,} its output is zero), the skip connection ensures that all original information is retained. 
Building on this concept, we designed the CNN-Adapter for adapting convolutional layers since more convolutional layers are involved at SAM 2 compared to SAM. The CNN-Adapter uses a point-wise convolutional layer to downsample the channel dimension, reducing complexity, followed by a depth-wise convolutional layer to capture spatial dimensions. Finally, a point-wise convolutional layer recovers the channel dimension to its original size. Inspired by ConvNext, we use only layer normalization and a GeLU activation function in this block. The bottleneck structure helps reduce complexity, and a parallel skip connection ensures that the output from the convolutional layers in SAM 2 is preserved. In the worst case, where the depth-wise layer learns nothing (\textit{i.e.,} its output is zero), the skip connection still retains all relevant information.

\section{Architecture of RFMedSAM 2}
Figure~\ref{fig:wholeArch_details}(a) illustrates the overall pipeline and architecture of RFMedSAM 2, which consists of three primary steps. In Step 0, an additional UNet model is employed to take medical images as input, generating initial multi-class mask predictions, which are then used to create auxiliary bounding boxes for the prompt requirements of SAM 2.
In Step 1, the medical images being input are involved into a modified image encoder to produce image embeddings, while the prompt encoder processes the auxiliary bounding boxes to generate point embeddings. These embeddings are passed to the modified mask decoder to generate masks and object pointers. The generated masks are then employed to create second bounding boxes for Step 2. A modified memory encoder processes both the generated masks and current frame features to produce memory features for the next step. Step 2 presents the second prediction by refining the initial predictions and performing the first refinement.
In Step 3, the same image features from the modified image encoder are input into a modified memory attention module, which establishes relationships with memory features from previous frames. The output from this memory attention mechanism is fed into the modified mask decoder, while the memory decoder also processes new point embeddings from the prompt encoder. Step 3 generates the third set of predictions and the second refinement, with the final mask prediction being output by the mask decoder.  
Figure~\ref{fig:wholeArch_details}(b)-(e) illustrates each component of RFMedSAM 2, described as follows. 

\subsection{Modified Image Encoder}
Figure~\ref{fig:wholeArch_details}(b) illustrates the redesigned image encoder. i) SAM works on natural images that have 3 channels for RGB while medical images have varied modalities as channels. There are gaps between the varied modalities of medical images and the RGB channels of natural images. 
Therefore, we design a sequence of two stacked convolutional layers to an invert-bottleneck architecture to learn the adaption from the varied modalities with any size to 3 channels. 
ii) SAM 2 employs Hiera~\cite{ryali2023hiera} that is hierarchical with multiscale output features as its image encoder backbone and a FPN module. Hiera consists of four stages with different feature resolutions and every stage contains various number of attention blocks. We insert our designed DWConvAdapter blocks into each attention block in Hiera. The output of each stage will be connected with one convolution in the FPN module. The latest output feature is up-sampled and summed with the second latest output feature as the image embedding. The third and fourth latest output feature are as skip connections to to incorporate high-resolution embeddings for the mask decoding. To adapt these convolutional layers, we insert our designed CNN-Adapters for the output features from the FPN module. 

\begin{figure*}[!t]
\centering
  \vspace{-0.2cm}
\includegraphics[width=1\linewidth]{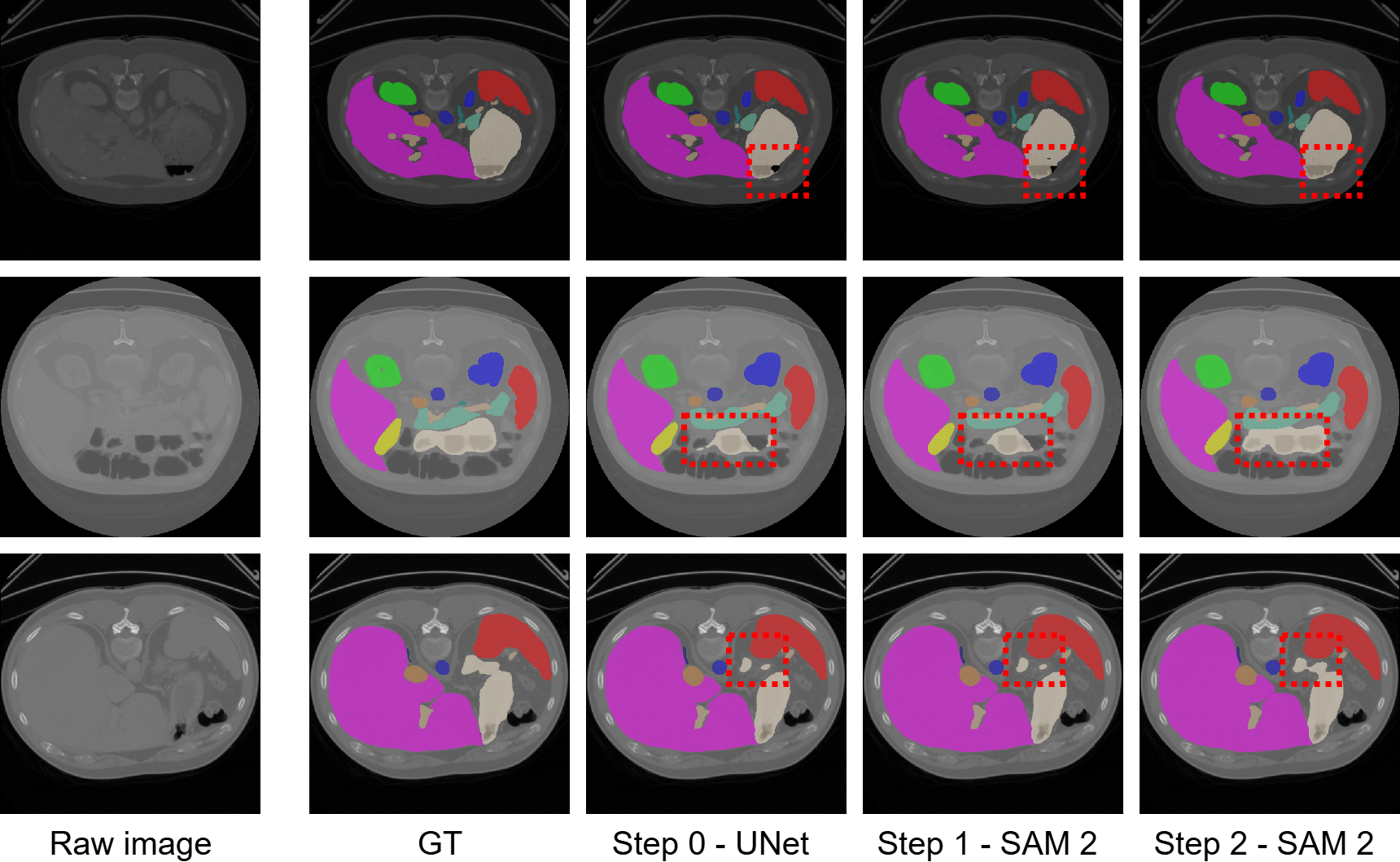}
      \caption{More visualization of two refinements.} 
\vspace{-0.2cm}
\label{fig:morerefinement}
\end{figure*}

\subsection{Modified Mask Encoder}
Figure~\ref{fig:wholeArch_details}(c) illustrates the redesigned mask encoder. The mask encoder contains two subsequent transformers and two following convolutional layers. i) Each transformer first applies self-attention to the prompt embedding. We insert an adapter behind the self-attention. Then, a cross-attention block is adopted for tokens attending to image embedding. We insert an adapter behind the cross-attention. Next, we insert a adapter parallel to an MLP block. Finally, a cross-attention block is utilized for image embedding attending to tokens. We insert a DWConvAdapter behind the cross-attention. In this way, our model can learn the spatial information for the image embedding and adapt information for the prompt embedding. 
ii) We inserted a CNN-Adapter behind the two following convolution layers to adapt convolutional layers from natural images to medical images. 

\subsection{UNet, Modified Memory Encoder and Modified Memory Attention.}
Figure~\ref{fig:wholeArch_details}(d) and (e) illustrate the UNet and the redesigned memory encoder, respectively. 
i) UNet is designed with a symmetrical encoder-decoder structure with skip connections. The encoder consists of several stages, each formed by a sequence of convolutional layers followed by down-sampling layers, progressively increasing the number of channels while reducing the spatial resolution to capture different deep-level features. The decoder upsamples the feature maps using transposed convolutions to restore spatial resolution and refine predictions. Skip connections between corresponding encoder and decoder layers enable the network to retain fine-grained spatial details, enhancing localization accuracy.  
ii) The memory encoder comprises two modules: the mask downsampler, which processes predicted masks, and the fuser, which integrates image features and mask features. To adapt these CNN-based modules to medical images, a CNN-Adapter is inserted after each module.
iii) The memory attention module stacks several transformer blocks, the first one taking the image encoding from the current frame as input. Each block performs self-attention, followed by cross-attention to memory features. Therefore, we inserted our designed DWConvAdapter blocks into each attention block since the transformer blocks process the image embedding with the spatial dimension.

\section{Impact of Auxiliary Losses on Image Encoder Parameter Updates if Prompt Generator Built with Image Encoder}
Figure~\ref{fig:motivate}(4d) illustrates a hierarchical structure with convolutional layers combined with multi-level features from the image encoder. The features with a lower resolution gradually increase the resolution by convolution layers and then combined with higher resolution features. Auxiliary loss functions are employed to supervise between the predicted masks and the ground truth. Although this approach achieves a DSC of 84.93\%, the result is not competitive. During training, both the auxiliary losses from the generated masks and the final output losses from SAM 2 influence the update of the image encoder parameters, which constitute a significant portion of the model. However, these two types of losses, due to their distinct architectural differences, are challenging to optimize simultaneously and achieve a balanced update for the image encoder parameters. 

\begin{figure}[!hbtp]
\centering
  \vspace{-0.2cm}
\includegraphics[width=1\linewidth]{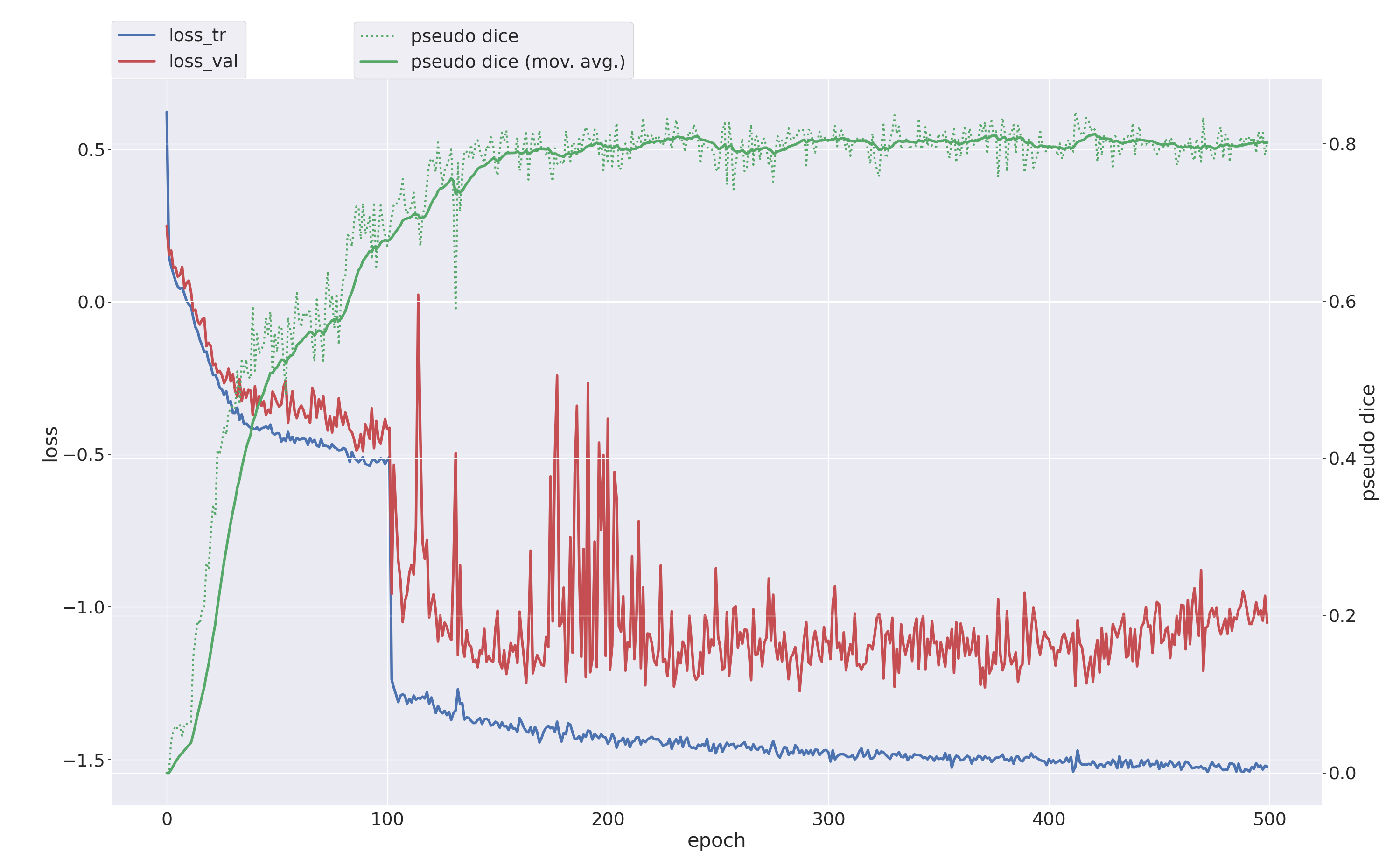}
      \caption{Oscillated losses if prompt generator built with image encoder.} 
\vspace{-0.2cm}
\label{fig:twolosses_1}
\end{figure}


We conduct experiments to validate the insights presented in Figure~\ref{fig:twolosses_1}. The training process is divided into two phases: one phase updates the parameters based solely on the auxiliary losses supervised by the auxiliary loss function, while the other phase updates all parameters based on both the auxiliary loss function and the final output loss function. The results indicate that after the second phase begins, the validation loss oscillates and is in an unstable state shown in the red line. The dice of the auxiliary masks present an unstable state since the final output losses affect the update of the image encoder and then affect the accuracy of the auxiliary masks. 

In conclusion, using a prompt generator built with the image encoder creates a challenge in balancing the update of the image encoder's parameters. As a result, we abandon this approach and instead employ an independent U-Net to generate masks and subsequently produce the corresponding bounding boxes. 


\section{Implementation Details}
We utilize some data augmentations such as rotation, scaling, Gaussian noise, Gaussian blur, brightness, and contrast adjustment, simulation of low resolution, gamma augmentation, and mirroring. We set the initial learning rate to 0.001 and employ a ``poly'' decay strategy in Eq.~\eqref{equa:polydecay}.
\begin{equation}
    lr(e)= init\_lr \times (1 - \frac{e}{\rm MAX\_EPOCH})^{0.9},
\label{equa:polydecay}
\end{equation}
where $e$ means the number of epochs, MAX\_EPOCH means the maximum of epochs, set it to 1000 and each epoch includes 250 iterations. We utilize SGD as our optimizer and set the momentum to 0.99. The weighted decay is set to 3e-5. We utilize both cross-entropy loss and dice loss by simply summing them up as the loss function. We utilize instance normalization as our normalization layer. we employ the deep supervision loss for the supervision of the U-Net. All experiments are conducted using two NVIDIA RTX A6000 GPUs with 48GB memory.

\noindent\textbf{Deep Supervision.} The U-Net network is trained with deep supervision. 
For each deep supervision output, we downsample the ground truth segmentation mask for the loss computation with each deep supervision output. The final training objective is the sum of all resolutions loss:
\begin{equation}
    \begin{aligned}
            \mathcal{L} = w_1 \cdot \mathcal{L}_1 + w_2 \cdot \mathcal{L}_2 + w_3 \cdot \mathcal{L}_3 + \cdot \cdot \cdot w_n \cdot \mathcal{L}_n
    \end{aligned}
    \label{equa:finalloss}
\end{equation}
where the weights halve with each decrease in resolution~(\textit{i.e.,} $w_2 = \frac{1}{2} \cdot w_1; w_3 = \frac{1}{4} \cdot w_1$, etc), and all weight are normalized to sum to 1. Meanwhile, the resolution of $\mathcal{L}_1$ is equal to $2 \cdot \mathcal{L}_2$ and $4 \cdot \mathcal{L}_3$.

\section{More Visualization of Two Refinements}
In Figure~\ref{fig:morerefinement}, we present additional qualitative results showcasing the refinements at different stages. With the two refinements, the results clearly illustrate the progressive improvement in segmentation accuracy, emphasizing the effectiveness of our model's refinement process.

\end{document}